\documentclass[conference]{IEEEtran}
\IEEEoverridecommandlockouts
\usepackage{makecell}
\usepackage{cite}
\usepackage{amsmath,amssymb,amsfonts}

\usepackage{graphicx}
\usepackage{textcomp}
\usepackage{xcolor}
\usepackage{algorithm}
\usepackage{algorithmic}
\usepackage{preambles}

\makeatletter
\newcommand{\linebreakand}{%
  \end{@IEEEauthorhalign}
  \hfill\mbox{}\par
  \mbox{}\hfill\begin{@IEEEauthorhalign}
}
\makeatother

\usepackage[caption=false,font=footnotesize]{subfig}
\def\BibTeX{{\rm B\kern-.05em{\sc i\kern-.025em b}\kern-.08em
    T\kern-.1667em\lower.7ex\hbox{E}\kern-.125emX}}
\begin{document}

\title{An Uncertainty-Weighted Decision Transformer for Navigation in Dense, Complex Driving Scenarios\\
{}
\thanks{*This work is supported by Carnegie Mellon University’s Safety21 National University Transportation Center, which is sponsored by the US Department of Transportation under grants 69A3552344811/69A3552348316.}% <-this % stops a space
\thanks{Some icons and graphics in this work were sourced from Vecteezy.com under the Free License.}
}

\author{\IEEEauthorblockN{Zhihao Zhang}
\IEEEauthorblockA{\textit{Electrical and Computer Engineering} \\
\textit{The Ohio State University}\\
Columbus, USA \\
zhang.11606@osu.edu}
\and
\IEEEauthorblockN{Chengyang Peng}
\IEEEauthorblockA{\textit{Mechanical and Aerospace Engineering} \\
\textit{The Ohio State University}\\
Columbus, USA \\
peng.947@osu.edu}
\and
\IEEEauthorblockN{Minghao Zhu}
\IEEEauthorblockA{\textit{Electrical and Computer Engineering} \\
\textit{The Ohio State University}\\
Columbus, USA \\
zhu.1385@osu.edu}
\linebreakand
\IEEEauthorblockN{Ekim Yurtsever}
\IEEEauthorblockA{\textit{Center for Automotive Research} \\
\textit{The Ohio State University}\\
Columbus, USA \\
yurtsever.2@osu.edu}
\and
\IEEEauthorblockN{Keith A. Redmill}
\IEEEauthorblockA{\textit{Electrical and Computer Engineering} \\
\textit{The Ohio State University}\\
Columbus, USA \\
redmill.1@osu.edu}
}

\maketitle

\begin{abstract}

%We present the Uncertainty-Weighted Decision Transformer, a sequence-modeling controller for autonomous driving in complex roundabout traffic.  The model encodes four-layer occupancy grids with a spatial attention block and captures long-horizon dynamics through a temporal transformer.  During offline training, token-level predictive entropy down-weights unreliable transitions, allowing the agent to focus on high-confidence experience and to learn robust policies from limited data.  Compared with a vanilla Decision Transformer and strong reinforcement-learning baselines, our method achieves higher accumulated reward, a greater exit success rate, and fewer collisions in multi-lane roundabout simulations.  These results show that coupling uncertainty estimation with transformer-based sequence learning improves safety and reliability in dynamic driving environments.

Autonomous driving in dense, dynamic environments requires decision-making systems that can exploit both spatial structure and long-horizon temporal dependencies while remaining robust to uncertainty. This work presents a novel framework that integrates multi-channel bird’s-eye-view occupancy grids with transformer-based sequence modeling for tactical driving in complex roundabout scenarios. To address the imbalance between frequent low-risk states and rare safety-critical decisions, we propose the Uncertainty-Weighted Decision Transformer (UWDT). UWDT employs a frozen teacher transformer to estimate per-token predictive entropy, which is then used as a weight in the student model’s loss function. This mechanism amplifies learning from uncertain, high-impact states while maintaining stability across common low-risk transitions. Experiments in a roundabout simulator, across varying traffic densities, show that UWDT consistently outperforms other baselines in terms of reward, collision rate, and behavioral stability. The results demonstrate that uncertainty-aware, spatial-temporal transformers can deliver safer and more efficient decision-making for autonomous driving in complex traffic environments.
\end{abstract}


\section{Introduction}
Autonomous driving is widely regarded as a key technology for improving road safety, reducing congestion, and increasing transportation efficiency. Achieving these benefits requires decision-making systems that can function reliably in dynamic, multi-agent environments such as intersections and roundabouts. In such settings, vehicles must perceive complex surroundings, anticipate interactive behaviors, and plan maneuvers that remain safe, efficient, and compliant with traffic rules even under uncertainty~\cite{yurtsever2020survey,kiran2021deep,treiber2000congested}.

Common approaches to handle autonomous driving decision-making can be broadly categorized into modular rule-based planners, imitation learning, search-based methods, and reinforcement learning (RL). Modular planners decompose the driving task into perception, prediction, behavior planning, and control modules, enabling interpretability but requiring extensive manual design and tuning by subject experts. Imitation learning, particularly behavior cloning, learns policies directly from expert demonstrations but suffers from covariate shift and a dependence on large, diverse datasets. Search-based planners, such as Monte Carlo Tree Search (MCTS), generate interpretable maneuvers by simulating and evaluating candidate action sequences, yet their computational cost scales poorly with traffic density, limiting real-time applicability~\cite{coulom2006efficient,kocsis2006bandit,coquelin2007bandit,hren2008optimistic}. RL methods offer the potential to learn adaptive strategies through interaction, but online RL involves unsafe exploration, and many studies focus on simplified scenarios with limited generalization~\cite{kiran2021deep,zhang2025extensive,peng2025reinforcement}.

Decision Transformers (DTs) have recently emerged as a promising approach for offline reinforcement learning, reframing decision-making as a conditional sequence modeling problem~\cite{chen2021decision,janner2021offline}. DTs leverage transformer architectures to capture long-horizon temporal dependencies without requiring unsafe online exploration. However, standard DTs are prone to performance degradation when encountering out-of-distribution states, often overfitting to frequent low-risk patterns while under-representing rare but safety-critical situations~\cite{guo2024leveraging}. In other domains, reweighting strategies based on predictive uncertainty are widely used, such as focal loss, entropy-based knowledge distillation, and uncertainty-aware RL~\cite{lin2017focal,chen2025uncertainty,yeung2022unified,li2023uncertainty,wu2021uncertainty,yi2024teaching}. These methods have proven effective in mitigating imbalance and improving robustness.

\begin{figure}[t]
    \centering
    \includegraphics[width=1\linewidth]{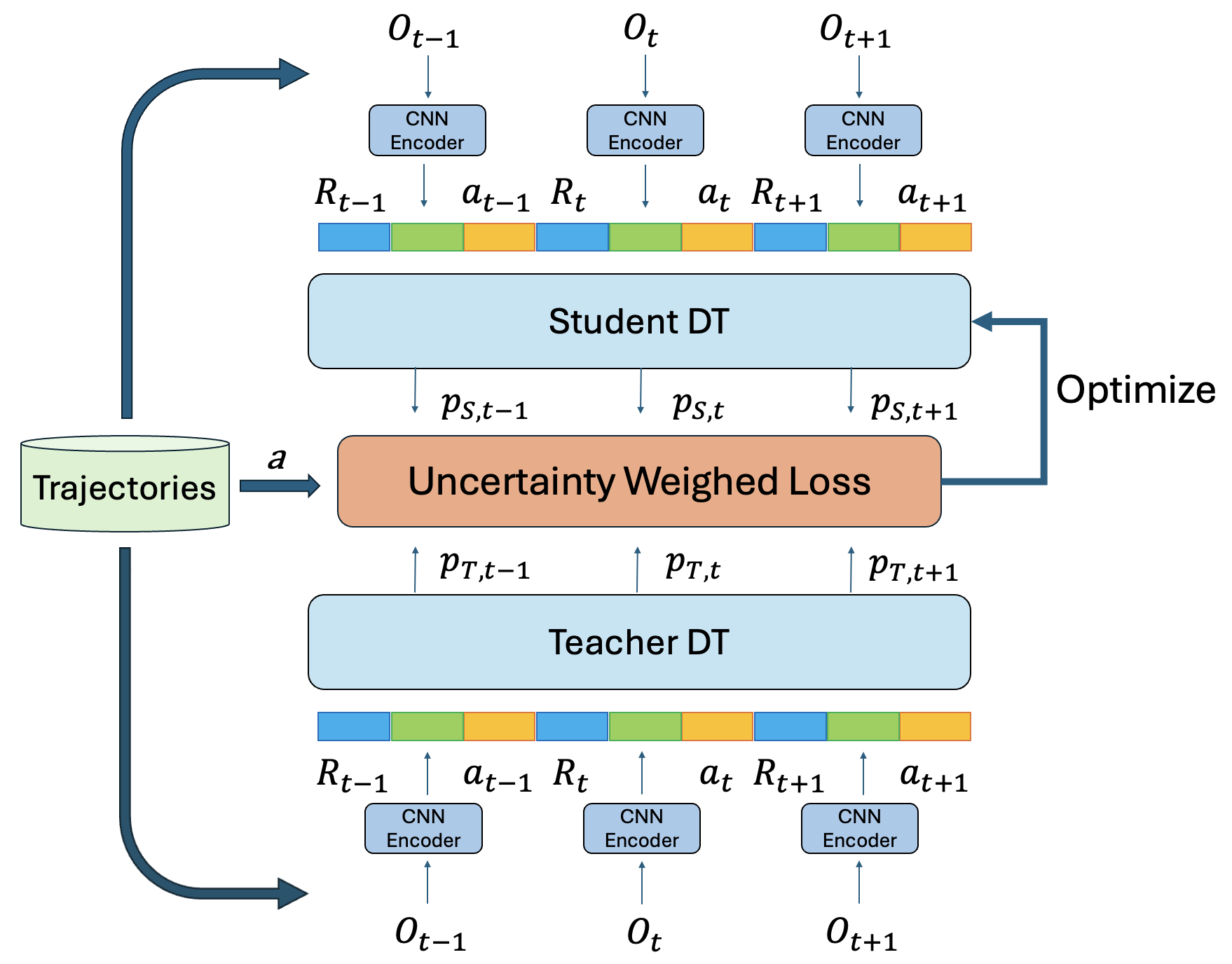}
    %alt=Automated vehicle control pipeline
    \caption{Teacher DT predictive uncertainty is estimated via the entropy of the action distribution. This entropy then acts as a weighting signal during student DT training, modulating the impact of each action prediction on the loss. }
    \label{HDT}
\end{figure}

Motivated by these insights, we propose the Uncertainty-weighted Decision Transformer (UWDT), a novel offline RL framework that combines multi-channel bird’s-eye-view occupancy grids for spatial representation with uncertainty-aware training (Figure~\ref{HDT}). A frozen teacher DT provides per-token predictive entropy, which is used to reweight the student DT’s loss, emphasizing rare, high-uncertainty states critical for safety. Our main contributions are:
\begin{enumerate}

\item An uncertainty-weighted training scheme that adaptively amplifies learning signals from safety-critical scenarios without altering the model architecture.
\item A comprehensive evaluation in a high-fidelity roundabout simulation across varying traffic densities, showing that UWDT outperforms baseline methods in reward, collision rate, and behavioral stability.
\end{enumerate}

The paper is organized as follows. Section~\ref{senario settings} defines the complex driving scenario. Section~\ref{problem formulation} introduces the reinforcement learning formulation for driving. Section~\ref{proposed method} presents our proposed method, and Section~\ref{results} provides the experimental results and analysis. Finally, Section~\ref{conclusion} concludes the paper and discusses future research directions.

\section{Scenario Settings}
\label{senario settings}
\begin{figure}[t]
    \centering
    \includegraphics[width=1\linewidth]{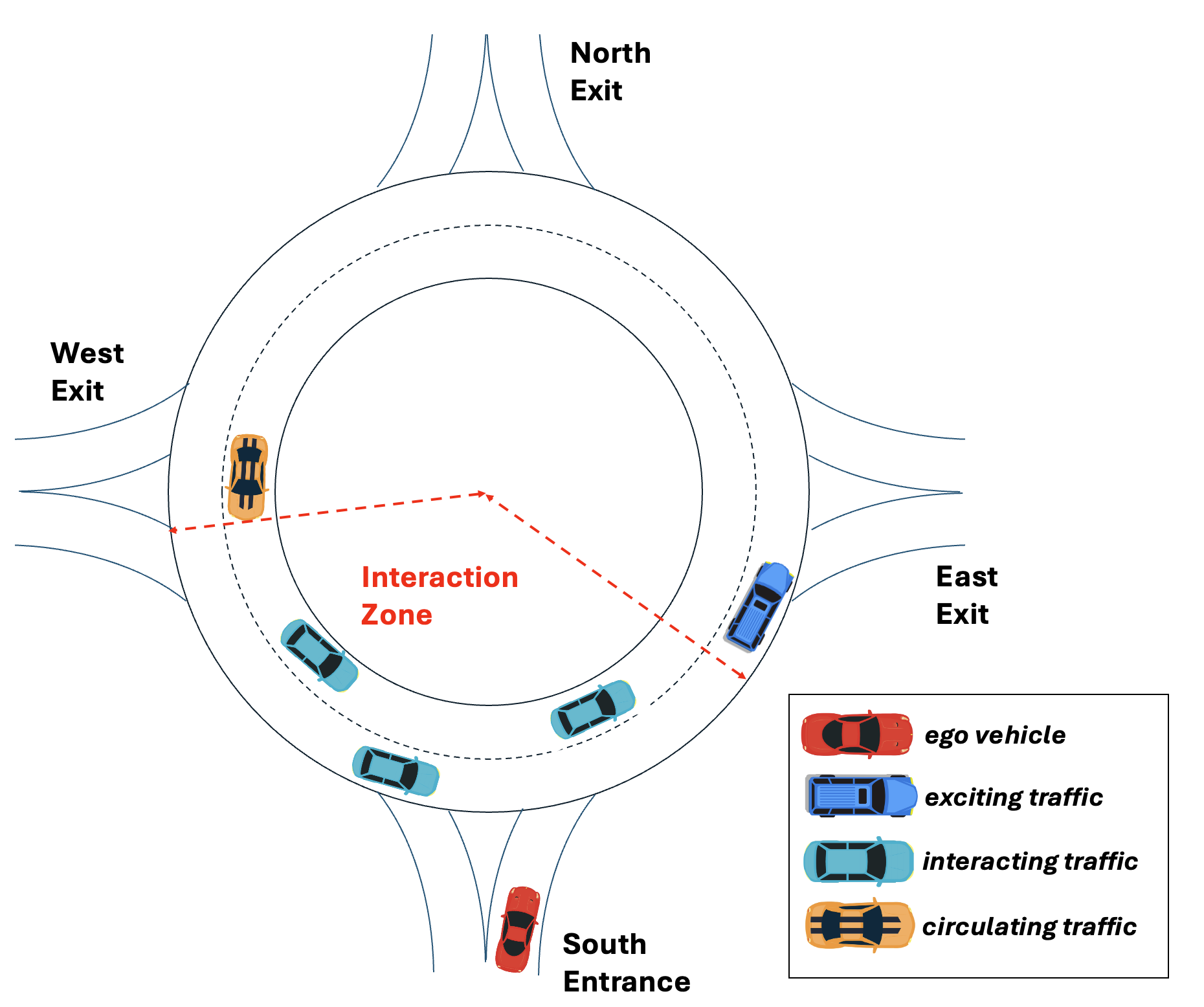}
    \caption{A four-arm, two-lane roundabout is used in our experiments. 
    The \textcolor{red}{ego vehicle} enters from the south, \textcolor{orange}{circulating traffic} is already inside the roundabout and travels behind interacting traffic, \textcolor{teal}{interacting traffic} travels immediately ahead and behind the ego, and \textcolor{blue}{exiting traffic} leaves from the east arm. 
    The dashed red polygon denotes the interaction zone, representing the ideal spatial region where traffic vehicles may interfere with or constrain the ego vehicle's entry into the roundabout.}
    \label{fig:roundabout}
\end{figure}

In our experimental setup, shown in Figure~
\ref{fig:roundabout}, the planner’s mission is to guide the ego vehicle entering the four arm, two lane roundabout from the south arm safely and efficiently to its fixed exit at the north arm, while negotiating circulating, interacting, and exiting traffic.

\subsection{Initial states and randomness}
The initial lateral and longitudinal positions for background vehicles are described below. For every background vehicle, the initial speed
\[
    v \sim \mathcal{N}\bigl(16,\,0.1^{2}\bigr)\,m/s 
\]
is sampled independently and a Gaussian perturbation ($\sigma\!=\!1\,\text{m}$) is added to each longitudinal position. Each vehicle perturbs the parameters of its Intelligent Driver Model (IDM) controller to introduce variation in acceleration and time-gap preferences. 
These stochastic elements ensure that every episode presents the ego vehicle with a unique configuration of circulating, interacting, and exiting traffic.

\subsection{Vehicle categories}
Traffic is divided into four functional groups that correspond to Figure~\ref{fig:roundabout}:

\textbf{Ego vehicle} (red). 
          Spawned $125\,\text{m}$ before the south entrance on the inbound lane with an initial speed of 8 m/s. 
          Its fixed goal is the exit at the north arm.
          
\textbf{Circulating traffic} (orange). 
      A random number of vehicles
      \[
          N_{\text{circ}}\sim\mathcal{U}\{0,1,2\}
      \]
      are initialized on the western approach and enter the roundabout, occupying either the inner or outer circulating lane. 
      Each vehicle selects a random destination among the north, east, or west exits, simulating diverse circulation behaviors.

\textbf{Interacting traffic} (teal). 
      A random number of exiting vehicles
      \[
          N_{\text{interact}} \sim \mathcal{U}\{0,1,2,3,4\}
      \]
      are inserted at varying positions along the circulating lanes. 
      These vehicles are likely to interfere with the ego vehicle’s entry decision, as they may block or yield at the merging point. 
      Their initial longitudinal offsets and lane assignments are randomized to reflect realistic variability in traffic flow.

\textbf{Exiting traffic} (blue). 
      Two vehicles are placed on the eastern approach, each at a distance of $50\,\text{m}$ upstream from the roundabout. 
      They proceed directly toward the eastern exit, simulating through-traffic that does not engage with the ego vehicle.

By increasing the number of interacting vehicles \(N_{\text{interact}}\), we systematically escalate planning difficulty: higher \(N_{\text{interact}}\) values reduce available gaps, raise uncertainty in gap acceptance, and force the planner to negotiate tighter merge windows or defer entry behaviors, all of which dramatically elevate roundabout entry complexity under uncertainty. This design allows us to exploit rising MCTS computational costs, decision latency, and predictive entropy as a quantitative metric for scenario complexity, attributable to higher interacting vehicle counts.

\section{Problem Formulation}
\label{problem formulation}
\subsection{Reinforcement Learning for Driving Decision-Making}
The driving decision-making problem for vehicles in high-speed environments can be modeled as a Markov Decision Process (MDP)~\cite{sutton2018reinforcement}
\begin{align}
\label{eq:mdp}
    \mathcal{M} = (\mathcal{S}, \mathcal{A}, \mathcal{P}, r, \gamma),
\end{align}
where $\mathcal{S}$ is the state space; $\mathcal{A}$ is the action space; $\mathcal{P}$ represents the state transition model; $r :\mathcal{S}\times\mathcal{A}\xrightarrow{}\mathbb{R}$ is a reward function and $\gamma \in (0,1]$ is a discount factor.

\newsec{State Space.}
We adopt the default observation from \textit{highway‑env}~\cite{highway-env}, a \(4 \times 41 \times 50\) Occupancy grid~\cite{hoermann2018dynamic,coue2006bayesian,danescu2011modeling,negre2014hybrid,rummelhard2015conditional}. 
The four channels correspond to \textit{presence}, longitudinal velocity \(v_x\), lateral velocity \(v_y\), and a binary \textit{on\_road} indicator. 
The grid is centered on the ego vehicle, spans \(100\,\text{m} \times 82\,\text{m}\), and is discretised at \(2\,\text{m}\) resolution. All features are clipped to predefined ranges and linearly scaled to \([-1, 1]\) to ensure numerical stability and compatibility with the CNN encoder used by our policy. This compact tensor representation preserves both spatial layout and dynamic context, enabling efficient convolutional processing of surrounding traffic interactions.

\newsec{Action Space.}
At each decision step $t$ the policy selects one action $a_t\in\mathcal{A}$ from
\[
\mathcal{A}=\{\text{llc},\ \text{rlc},\ \text{acc},\ \text{dec},\ \text{cruise}\}.
\]
Here, \text{llc} stands for left lane change; \text{rlc} stands for right lane change; \text{acc} stands for acceleration; \text{dec} stands for deceleration; \text{cruise} stands for maintain speed.
This high-level action is then translated by a low-level motion planner into continuous control signals \((a_{acc},\theta_{lat} )\) for acceleration and steering, where $a_{acc} \in [-1, 1] \mathrm{m/s^2}$ and $\theta_{lat}\in[-\frac{\pi}{36},\ \frac{\pi}{36}]\,\mathrm{rad}$.

\newsec{Reward Function.}
Following the reward structures adopted in recent tactical-driving studies~\cite{leurent2019social,wang2024reachability}, the objective of the roundabout driving task encourages fast and safe driving while discouraging unnecessary lane changes.

Three scalar reward components are derived as follows:

\begin{itemize}
\item The collision indicator $c_t$ equals $1$ if the ego-vehicle has crashed and $0$ otherwise.
\item The speed reward indicator $v_t$ equals $1$ if the ego-vehicle’s speed is within the range $[8,16]~\text{m/s}$, and $0$ otherwise.
\item The lane-change indicator $\ell_t$ equals $1$ when the chosen action commands a lane change, and $0$ otherwise.
\end{itemize}

The raw reward at each time step combines these components linearly with fixed weights from the default configuration:
\begin{equation}
r_t^{\text{raw}} = w_c c_t + w_v v_t + w_{\ell} \ell_t
\end{equation}
where the collision weight $w_c=-1$ imposes a severe penalty upon collisions, the speed weight $w_v=0.2$ provides a modest reward for maintaining high velocities (favoring efficient traffic flow without reckless acceleration), and the lane-change weight $w_{\ell}=-0.05$ penalizes unnecessary lateral movements to encourage stable driving patterns. The reward is linearly scaled to $[0,1]$ as follows:
\begin{equation}
\tilde{r}_t = \frac{r_t^{\text{raw}} - (w_c+w_{\ell})}{w_v -  (w_c+w_{\ell})},
\end{equation}
such that $\tilde{r}_t=0$ at collision and $\tilde{r}_t=1$ when driving at the maximum speed without lane changes. This ensures a consistent reward range for training across episodes.

\subsection{Training Dataset}
To construct a high-quality offline training dataset, we employ MCTS~\cite{kocsis2006bandit} to generate 5,400 expert roundabout roll-outs. By leveraging the reward function, the trajectories generated by MCTS are well aligned with the downstream RL objective. Compared with heuristic or human-driven demonstrations, this procedure yields near-optimal yet unbiased expert trajectories that provide consistent and high-quality supervision for our DT and Transformer based behavior cloning baselines.

% Building on this corpus, we introduce the Uncertainty-Weighted Decision Transformer (UWDT).  A frozen, high-capacity teacher provides token-level predictive entropy that acts as an automatic importance weight, increasing gradient focus on ambiguous yet safety-critical frames that are statistically rare in the expert data.  The student shares the vanilla DT architecture but is optimized with this entropy-scaled loss, simultaneously addressing class imbalance and covariate shift without architectural change.  Compared with value-based offline RL or straightforward behavior cloning, UWDT offers a principled route to risk-aware planning in a purely offline setting.

\section{Proposed method}
\label{proposed method}
In this section, we introduce UWDT, an offline Decision Transformer framework designed to improve robustness in rare and safety-critical traffic scenarios by reweighting the training objective. UWDT proceeds in three stages:

% Offline decision‑making policies based on the DT formulation achieve state‑of‑the‑art performance by reframing RL as conditional sequence modeling over returns, states, and actions~\cite{chen2021decision}.  
% During training, every transition receives the same loss weight. This uniform weighting causes DTs to memorize abundant, low‑risk regimes while leaving rare yet safety‑critical states (e.g., roundabout entries, forced merges, and short‑gap yields) insufficiently trained, even though these sparse cases dominate safety metrics~\cite{rodegerdts2010roundabouts}.

\begin{enumerate}
\item \textbf{Teacher fitting:} train $\pi_T$ on offline data and freeze its parameters.
\item \textbf{Entropy extraction:} run $\pi_T$ over the dataset to compute per-token entropies $H_t$ and normalize them into weights.
\item \textbf{Student distillation:} optimize $\pi_S$ with a weighted cross-entropy, allocating more gradient to high-entropy tokens.
\end{enumerate}
\noindent This procedure requires no architectural changes and unifies uncertainty-aware distillation with offline-RL weighting. By up-weighting uncertain, safety-critical states while preserving temporal context, UWDT improves robustness to rare driving scenarios.

We use a CNN encoder structure to capture spatial information. Each state tensor $s_t$ is encoded by a convolutional network:
three $3\times3$ convolutions with stride $2$ (channels $4\!\to\!32\!\to\!64\!\to\!128$),
ReLU activations, batch normalization, and spatial dropout.
The final feature map is flattened and projected to a fixed $d_e$--dimensional embedding $\phi_t = \phi(s_t) \in \mathbb{R}^{d_e}, d_e = 32$ via a lazy fully connected layer.

\subsection{Teacher Decision Transformer}

We adopt a standard DT as the teacher model.  Given an offline dataset 
$\mathcal{D}=\{\tau_i\}_{i=1}^N$ of trajectories $\tau=(s_1,a_1,r_1,\dots,s_T,a_T,r_T)$,
the return-to-go at time $t$ is
\begin{equation}
\label{eq:rtg}
R_t = \sum_{k=t}^{T}\gamma^{\,k-t}\,r_k
\end{equation}
with discount $\gamma\in[0,1)$.
At each step $t$, the input token context is the interleaved sequence
\[
X_{1:t} = \bigl(R_{t},\,s_{t},\,a_{t-1},\,R_{t-1},\,s_{t-1},\,a_{t-2},\,\dots\bigr).
\]
A causal Transformer $\pi_T(\cdot\mid X_{1:t};\theta_T)$ is trained  to predict the next action using input tokens.
Let $\mathcal{B}_{\mathrm{val}}$ denote the index set of valid tokens in a mini-batch and $M=|\mathcal{B}_{\mathrm{val}}|$.
Define the Teacher action distribution by $p_{T,t}(a)=\pi_T(a\mid X_{1:t};\theta_T)$.
The teacher loss is
\begin{equation}
\label{eq:teacher-loss}
\mathcal{L}_{T}(\theta_T)
= - \frac{1}{M}
\sum_{t \in \mathcal{B}_{\mathrm{val}}}
\log p_{T,t}(a_t).
\end{equation}
Gradient updates are performed with AdamW and a running‑mean‑square layer‑norm, following~\cite{chen2021decision}.

At test time, the next action is selected either by stochastic sampling or a greedy approach:
\[
a_t =
\begin{cases}
\mathrm{sample}\!\bigl(p_{T,t}(\cdot)\bigr), & \text{stochastic rollout}, \\[4pt]
\displaystyle \arg\max_{a}\, p_{T,t}(a), & \text{greedy rollout}.
\end{cases}
\]
In our experiments we adopt greedy rollout for stability and performance.
After convergence, $\theta_T$ is frozen and reused to compute per-token uncertainty for student training. Table~\ref{table:DTparam} lists the training parameters for the baseline DT and UWDT.

\begin{table}[H]
\centering
\begin{tabular}{l l l} \hline
Symbol & Meaning & Value \\ \hline
$K$                    & Context length (seq.\ window)               & 20 \\
$d$                    & Embedding dimension                         & 32 \\
$\gamma$               & Discount factor for returns‑to‑go           & 0.99 \\
$\eta$                 & Learning rate                               & $1\times10^{-5}$ \\
$B$                    & Batch size                      & 16 \\
$E$                    & Training epochs                             & 20 \\
$\lambda_{\text{wd}}$  & Weight decay (AdamW)                        & $5\times10^{-5}$ \\
$\rho_{\text{wu}}$     & Linear warm‑up ratio                        & 0.1 \\
$g_{\text{max}}$       & Gradient‑norm clip                          & 0.25 \\
$n_{\text{layer}}$     & Transformer decoder layers                  & 4 \\
$n_{\text{head}}$      & Attention heads                             & 1  \\ \hline
\end{tabular}
\vspace{5mm}
\caption{Backbone DT hyper‑parameters}
\label{table:DTparam}
\vspace{-4mm}
\end{table}

\subsection{Student Decision Transformer}

The student policy $\pi_S$ shares the architecture with the Teacher DT. After obtaining the trained Teacher DT, we estimate predictive uncertainty by computing the entropy of its action distribution. During student training, this entropy serves as an uncertainty-based weight $\bar{w}_t$, modulating the loss contribution of each training token. Compared to the Teacher DT, the student is architecturally identical but trained with entropy-weighted supervision.

Define the student action distribution by
$p_{S,t}(a)=\pi_S(a\mid X_{1:t};\theta_S)$.
For each valid token we compute the masked cross-entropy scaled by $\bar{w}_t$:
\begin{equation}
\label{eq:student-loss}
\mathcal{L}_{S}
= - \frac{1}{|\mathcal{B}_{\mathrm{val}}|}
\sum_{t \in \mathcal{B}_{\mathrm{val}}}
\bar{w}_t \,
\log p_{S,t}(a_t).
\end{equation}
Tokens with higher Teacher entropy carry larger gradient weight, encouraging the student to allocate additional capacity to states the Teacher found ambiguous. The Teacher’s predictive entropy is
\begin{equation}
\label{eq:entropy}
H_t
= - \sum_{a=0}^{A-1} p_{T,t}(a) \log p_{T,t}(a),
\qquad
0 \le H_t \le \log A .
\end{equation}
where $p_{T,t}(a)$ is the frozen Teacher action distribution and $A$ is the number of discrete actions.

Let $r>1$ denote the desired ratio between the largest and smallest per-token weights. We adopt a power mapping
\begin{equation}
\label{eq:weight-raw}
\tilde{w}_t = H_t^{\beta},
\end{equation}
which increases its weight monotonically with entropy, so tokens where the teacher is less certain (larger $H_t$) receive larger weights.

\begin{algorithm}[!t]
\caption{UWDT training }
\label{alg:uwdt}
\begin{algorithmic}[1]
\REQUIRE Frozen $\pi_T$, initial $\pi_S$; $r$, $w_{\max}$, $\eta$; $H_{\min},H_{\max}$
\STATE $\beta \gets \dfrac{\ln r}{\ln(H_{\max}/H_{\min})}$
\FOR{$\text{epoch}=1,\dots,E_S$}
  \STATE Sample mini-batch $\mathcal{B}\subset\mathcal{D}$; build $X_{1:t}$; get $\mathcal{B}_{\mathrm{val}}$; $M\gets|\mathcal{B}_{\mathrm{val}}|$
  \FORALL{$t\in\mathcal{B}_{\mathrm{val}}$}
    \STATE $p_{T,t}(a)\gets\pi_T(a\mid X_{1:t})$
    \STATE $H_t\gets-\sum_a p_{T,t}(a)\log p_{T,t}(a)$
    \STATE $\tilde{w}_t \gets H_t^{\beta}$
  \ENDFOR
  \STATE $\mu \gets \dfrac{1}{M}\sum_{j\in\mathcal{B}_{\mathrm{val}}}\tilde{w}_j$
  \FORALL{$t\in\mathcal{B}_{\mathrm{val}}$}
    \STATE $w_t \gets \tilde{w}_t / \mu$
    \STATE $\bar{w}_t \gets \min\{w_t,\,w_{\max}\}$
    \STATE $p_{S,t}(a)\gets\pi_S(a\mid X_{1:t})$
  \ENDFOR
  \STATE $\mathcal{L}_S \gets -\dfrac{1}{M}\sum_{t\in\mathcal{B}_{\mathrm{val}}}\bar{w}_t\log p_{S,t}(a_t)$
  \STATE $\theta_S \gets \theta_S - \eta\,\nabla_{\theta_S}\mathcal{L}_S$
\ENDFOR
\STATE \textbf{return} $\pi_S$
\end{algorithmic}
\end{algorithm}

\begin{table*}[!t]
\caption{Performance comparison on the roundabout task. }
\centering
\resizebox{0.95\textwidth}{!}{
\begin{tabular}{l c c c c c c c c}
\hline
Method &
\thead{Accumulated\\Reward} &
\thead{Average\\Speed [m/s]} &
\thead{Episode\\Length [s]} &
\thead{Travel\\Distance [m]} &
\thead{Reach Exit\\Rate [\%]} &
\thead{Collision\\Rate [\%]} &
\thead{Time--to--Exit\\$[\mathrm{s}]$} &
\thead{Halt Duration\\($v<1\,\mathrm{m/s}$) [s]} \\
\hline
CQL               & $12.82 \pm 7.55$ & $7.67 \pm 3.27$ & $14.60 \pm 7.82$ & $111.96 \pm 76.64$ & $14.00 \pm 34.74$ & $49.00 \pm 50.05$ & $21.44 \pm 1.62$ & $2.48 \pm 5.11$ \\[5pt]
SAC               & $20.66 \pm 1.99$ & $10.42 \pm 6.61$ & $21.71 \pm 1.27$ & $228.24 \pm 145.33$& $60.00 \pm 49.05$ & $1.50 \pm 7.06$  & $17.80 \pm 3.43$ & $6.01 \pm 7.76$ \\[5pt]
BC Transformer    & $13.63 \pm 7.82$ & $7.89 \pm 0.16$ & $15.26 \pm 8.06$ & $120.47 \pm 63.64$ & $0.00 \pm 0.00$ & $41.25 \pm 49.29$ & $22.00 \pm 0.00$ & $0.00 \pm 0.00$ \\[5pt]
DT                & $20.88 \pm 4.17$ & $15.03 \pm 1.73$ & $21.02 \pm 3.89$ & $315.86 \pm 68.81$ & $94.00 \pm 23.78$ & $6.00 \pm 23.78$ & $15.77 \pm 1.65$ & $0.00 \pm 0.00$ \\[5pt]
\textbf{UW-DT} & $\mathbf{21.69 \pm 2.01}$ & $\mathbf{15.31 \pm 0.79}$ & $\mathbf{21.79 \pm 1.90}$ & $\mathbf{333.55 \pm 33.73}$ & $\mathbf{98.75 \pm 11.12}$ & $\mathbf{1.25 \pm 11.12}$ & $\mathbf{15.52 \pm 0.98}$ & $\mathbf{0.00 \pm 0.00}$ \\
\hline
\end{tabular}}
\label{table:rewardcomparison}
\end{table*}

The exponent $\beta>0$ is chosen so that the dynamic range matches $r$:
\begin{equation}
\label{eq:gamma-from-r}
\beta
= \frac{\ln r}{\ln\!\big(H_{\max}/H_{\min}\big)}.
\end{equation}
Here \(H_{\min}\) and \(H_{\max}\) are the minimum and maximum teacher entropies measured over evaluation rollouts. In practice, we deploy trained Teacher DT for 400 episodes in the simulated environment and record the minimum and maximum entropy across all timesteps to quantify the teacher’s confidence range under deployment dynamics.

Because the scale of $\tilde{w}_t$ influences optimization, we normalize in each mini-batch:
\begin{equation}
\label{eq:weight-norm}
w_t = \frac{\tilde{w}_t}{ \frac{1}{M} \sum_{j=1}^{M} \tilde{w}_j },
\end{equation}
which sets the batch-mean weight to one and preserves the overall loss scale. Finally, we clip by a ceiling $w_{\max}$:
\begin{equation}
\label{eq:weight-clip}
\bar{w}_t = \min\{ w_t , w_{\max} \}.
\end{equation}
This limits extreme weights to prevent training instability. 

In practice, We set $w_{\max}=1.5$ and $r=1.3$.

\section{Results}
\label{results}

The environment is implemented with the \texttt{roundabout-v0} task in \textit{highway-env}~\cite{highway-env}. 
It models a single roundabout whose inner and outer circulating lanes have radii of $20\,\text{m}$ and $24\,\text{m}$, respectively. Simulation dynamics are integrated at $15\,\text{Hz}$, while the agent selects an action at $2\,\text{Hz}$; thus each episode lasts $11\,\text{s}$ or terminates upon collision. All experiments were conducted on an NVIDIA GeForce RTX 3070 Ti (GPU) and an AMD Ryzen 7 5800X (CPU).

\subsection{Performance Analysis}

In this section, we evaluate our approach under varying traffic densities and compare it against several offline RL and imitation-learning baselines. We consider four RL algorithms: Conservative Q-Learning (CQL), Soft Actor-Critic (SAC), DT and our proposed UWDT. For imitation learning, we include a Transformer-based Behavior Cloning (BC Transformer) model. Throughout this section, single-step rewards are normalized to $[0,1]$ and velocities are expressed in m/s.

\newsec{CQL:} Offline RL learns from fixed data. To counter overestimation from distribution shift, CQL trains a conservative Q-function that lower-bounds the policy value and adds a regularizer penalizing out-of-distribution actions.

\newsec{SAC:} SAC is an off-policy actor–critic algorithm based on the maximum-entropy framework. The policy optimizes expected return plus an entropy bonus, promoting better exploration and training stability across tasks.

\newsec{BC Transformer:} BC treats policy learning as supervised mapping from states to actions. It is implemented with a causal Transformer. 

\newsec{DT:} DT casts RL as sequence modeling: a causal Transformer is trained on sequences of returns, states, and actions with an autoregressive loss, avoiding unstable bootstrapping and enabling diverse behaviors by conditioning on desired returns.

\begin{table}[!b]
\caption{Action-distribution entropy statistics over testing}
\centering
\resizebox{0.4\textwidth}{!}{
\begin{tabular}{l c c c}
\hline
Method & \thead{Min entropy} & \thead{Max entropy} & \thead{Average entropy} \\
\hline
DT               & $1.14$ & $1.47$ & $1.15 \pm 0.02$ \\[5pt]
\textbf{UWDT} & $1.14$ & $\mathbf{1.34}$ & $\mathbf{1.15 \pm 0.01}$ \\
\hline
\end{tabular}
}

\label{table:entropycomparision}
\end{table}

Table~\ref{table:rewardcomparison} summarizes key performance metrics for the compared methods. The Episode Length measures the average duration until episode termination, capped at 22 timesteps unless a collision occurs earlier. Time-to-Exit denotes the average time taken to reach the exit; if a collision occurs first, the maximum duration of 22 is recorded. Halt Duration indicates time spent at speeds below \(1\,\mathrm{m/s}\).

CQL and BC rely purely on offline data. CQL penalizes uncertain, out-of-distribution actions, obtaining the lowest average reward of 12.82, the shortest trajectories of 14.60\,s, and an exit rate of 14\,\%. BC Transformer also struggles due to compounding errors and the lack of uncertainty handling, resulting in a low average reward of 13.63 and never reaching the exit. Its performance gap versus DT highlights poor generalization when the policy is purely imitative. SAC, an online RL method, yields an average reward of 20.66, comparable to DT and UWDT due to reward normalization during training. However, higher rewards mainly indicate fewer collisions rather than optimal driving. With an episode length of 21.71\,s and an exit success rate of 60\,\%, SAC often behaves overly cautiously, on average halting for 6.01\,s. Observations show that SAC waits excessively at the roundabout entrance, causing potential congestion despite clear merging opportunities. DT, which conditions actions on target returns, achieves notable improvements: an average speed of 15.03\,m/s, a travel distance of 315.86\,m, and an exit rate of 94\,\%. Yet the lack of explicit uncertainty awareness sometimes causes premature terminations due to risky actions. Its max entropy of 1.47 reflects uncertainty without a mechanism to exploit it for safer decisions.

The proposed UWDT explicitly integrates epistemic uncertainty, outperforming all baselines. It attains the highest average reward of 21.69, the fastest speed of 15.31\,m/s, the greatest travel distance of 333.55\,m, and a near-perfect exit rate of 98.75\,\%. Table~\ref{table:entropycomparision} corroborates these findings, showing that UWDT maintains a tightly controlled entropy range of 1.14--1.34, confirming better-calibrated uncertainty than DT. 
% \begin{figure*}[!t]
%   \centering

%   % Row 1 (a–c): velocity
%   \subfloat[]{\includegraphics[width=0.32\textwidth]{figures/5velocityvehiclenumber2.png}\label{fig:n2_velocities}}\hfill
%   \subfloat[]{\includegraphics[width=0.32\textwidth]{figures/5velocityvehiclenumber3.png}\label{fig:n3_velocities}}\hfill
%   \subfloat[]{\includegraphics[width=0.32\textwidth]{figures/5velocityvehiclenumber4.png}\label{fig:n4_velocities}}\\[0.5em]

%   % Row 2 (d–f): reward
%   \subfloat[]{\includegraphics[width=0.32\textwidth]{figures/5Rewardvehiclenumber2.png}\label{fig:n2_rewards}}\hfill
%   \subfloat[]{\includegraphics[width=0.32\textwidth]{figures/5Rewardvehiclenumber3.png}\label{fig:n3_rewards}}\hfill
%   \subfloat[]{\includegraphics[width=0.32\textwidth]{figures/5Rewardvehiclenumber4.png}\label{fig:n4_rewards}}

%   \caption{
%     Velocity profile (top row; a–c) and reward profile (bottom row; d–f) averaged over 20 episodes at three traffic densities.
%     (a,d) interacting-vehicle count uniformly drawn from \([0,2]\);
%     (b,e) three interacting vehicles;
%     (c,f) four interacting vehicles.
%     UWDT and DT maintain higher reward and speed; CQL and BC deteriorate as density increases; SAC shows gradual recovery.
%   }
%   \label{fig:reward_velocity_grid}
% \end{figure*}

\begin{figure*}[!t]
  \centering
    \includegraphics[trim={0cm 17.1cm 0cm 0cm},clip,width=\textwidth]{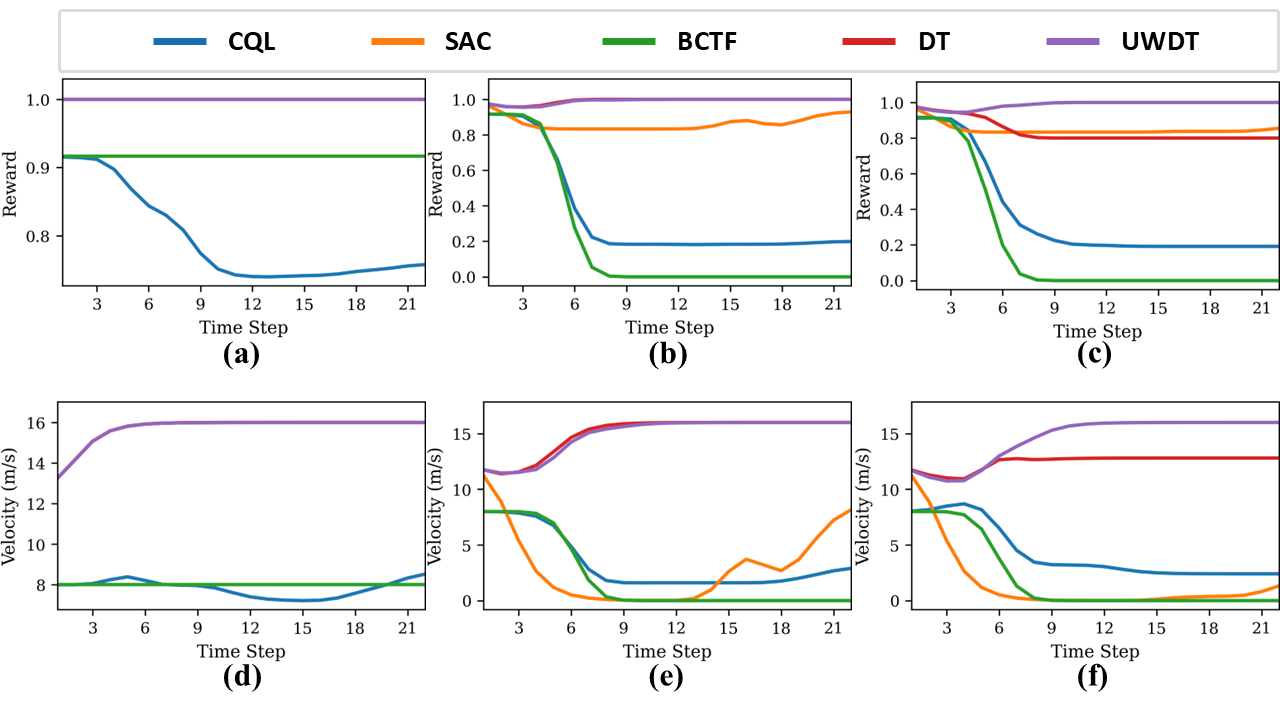}
             
  % Row 1 (a–c): velocity
  \subfloat[]{\includegraphics[width=0.33\textwidth]{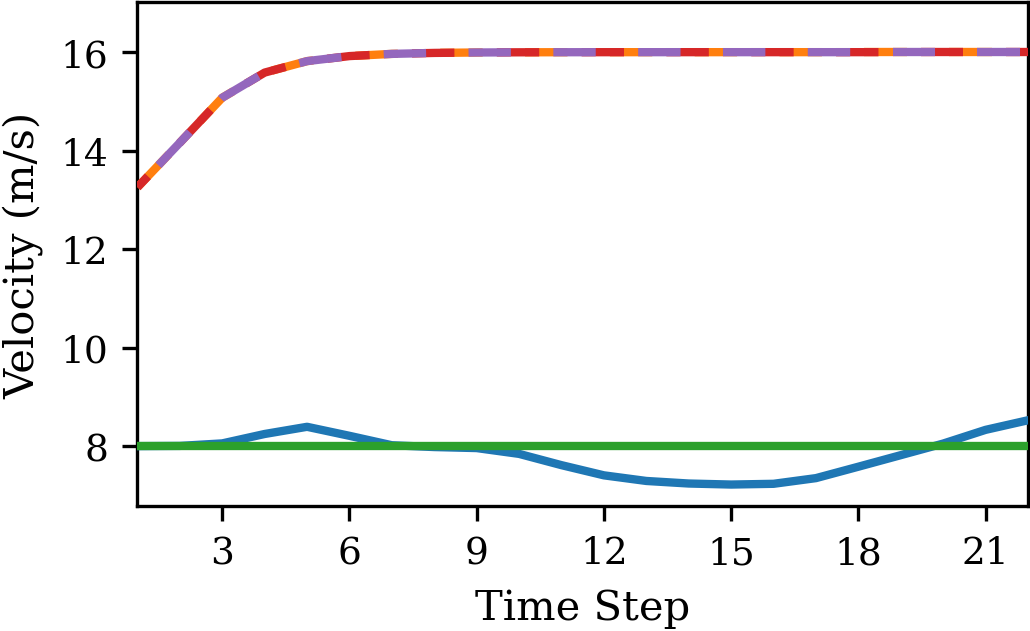}\label{fig:n2_velocities}}\hfill
  \subfloat[]{\includegraphics[width=0.33\textwidth]{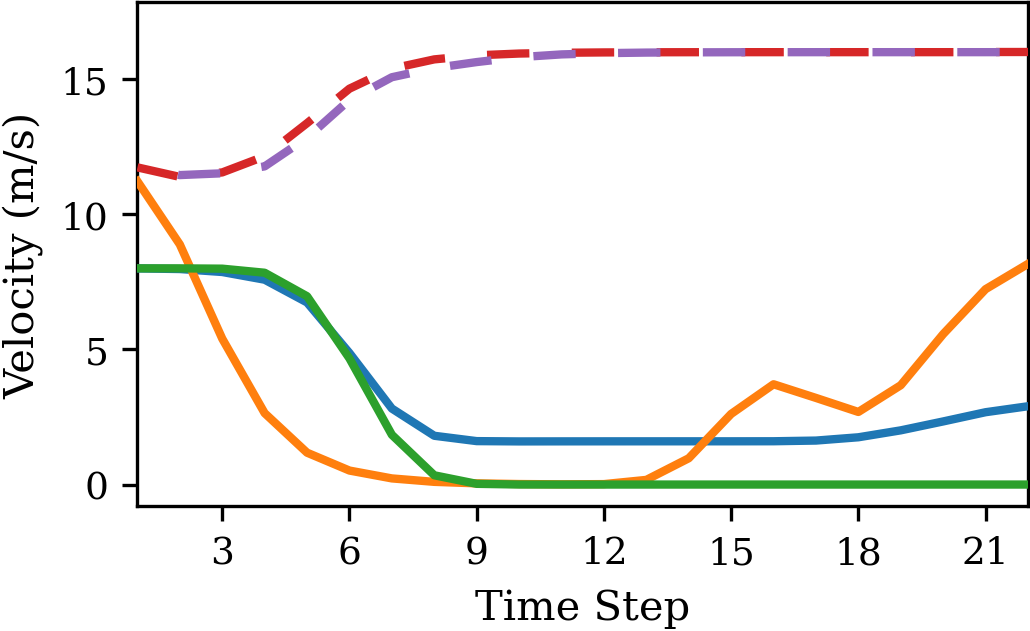}\label{fig:n3_velocities}}\hfill
  \subfloat[]{\includegraphics[width=0.33\textwidth]{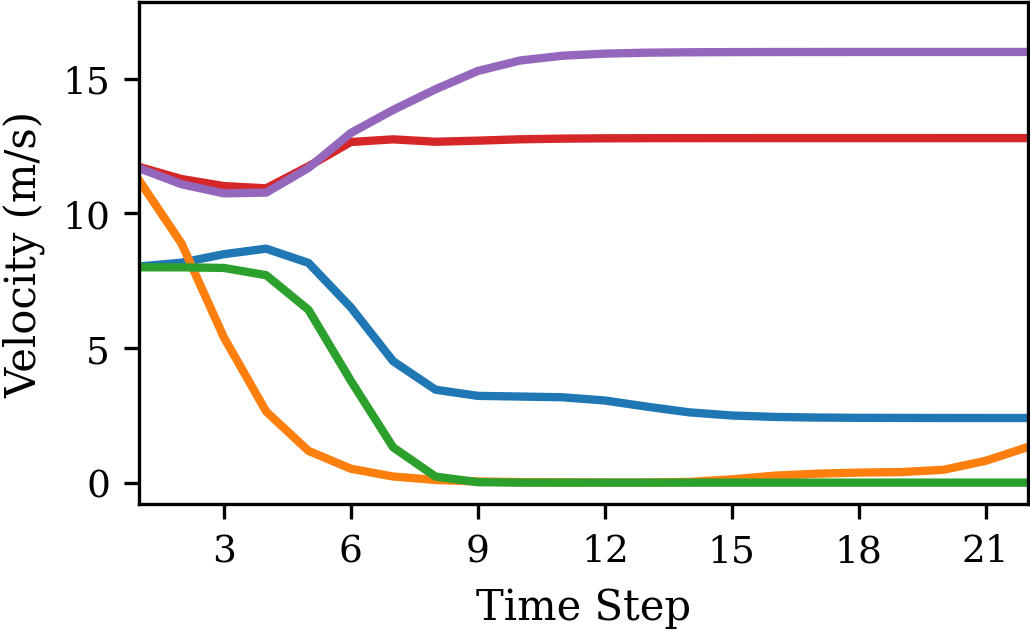}\label{fig:n4_velocities}}\\[0.5em]

  % Row 2 (d–f): reward
  \subfloat[]{\includegraphics[width=0.33\textwidth]{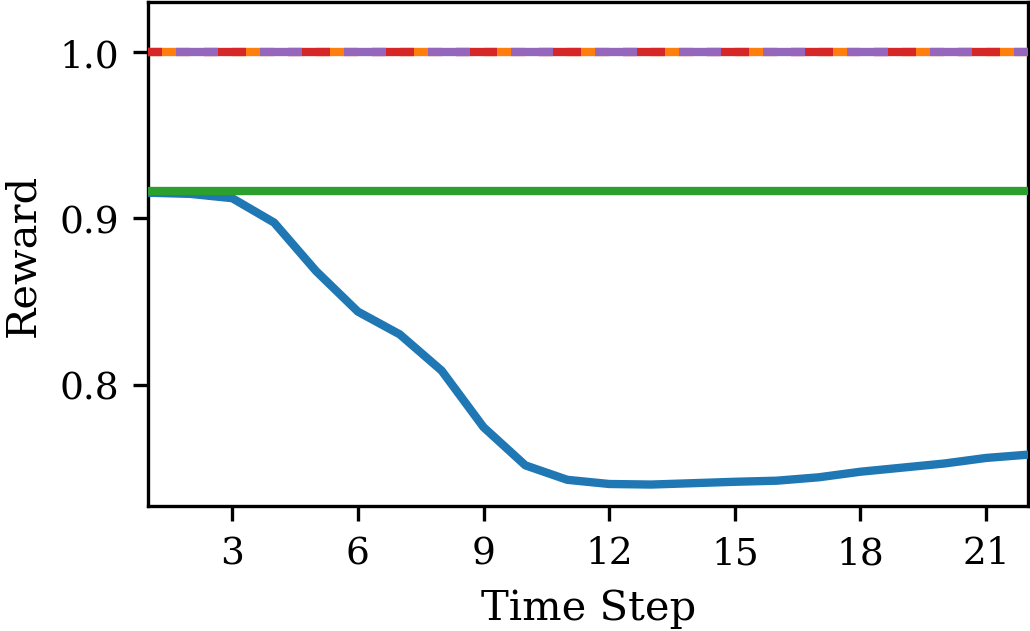}\label{fig:n2_rewards}}\hfill
  \subfloat[]{\includegraphics[width=0.33\textwidth]{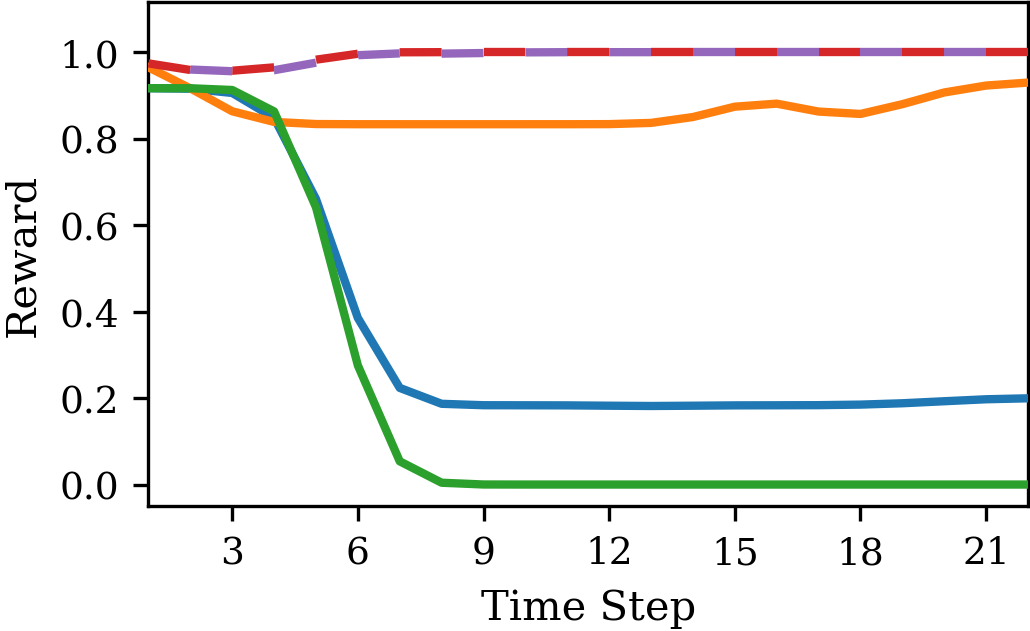}\label{fig:n3_rewards}}\hfill
  \subfloat[]{\includegraphics[width=0.33\textwidth]{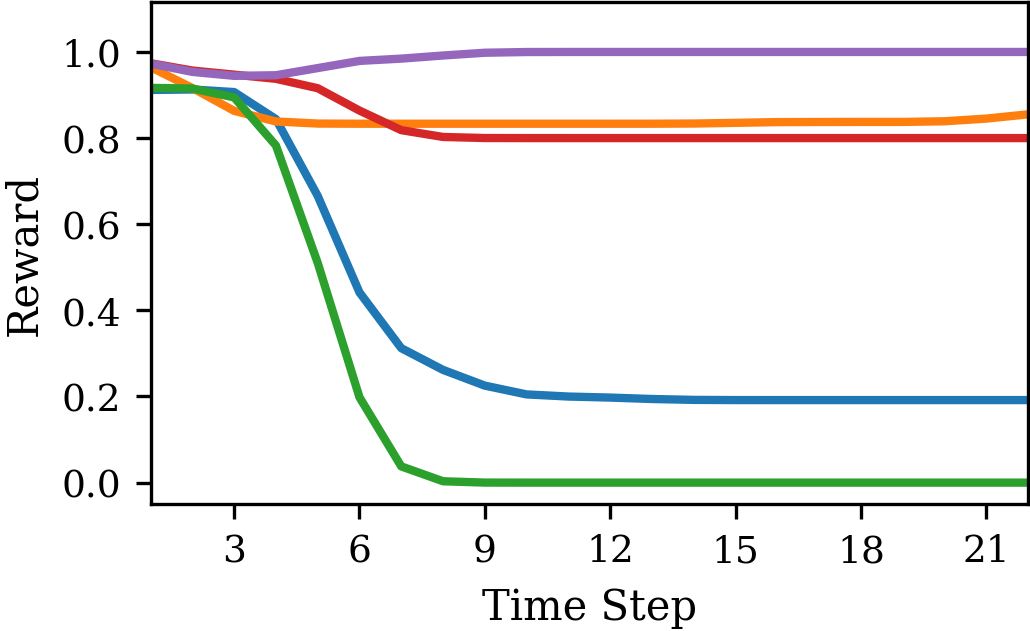}\label{fig:n4_rewards}}

  \caption{
    Velocity profile (top row; a–c) and reward profile (bottom row; d–f) averaged over 20 episodes at three traffic densities.
    In (a,d) the interacting-vehicle count is uniformly drawn from \([0,2]\);
    (b,e) involves three interacting vehicles; and
    (c,f) involves four interacting vehicles.
    UWDT and DT maintain higher reward and speed; CQL and BC deteriorate as density increases; SAC shows gradual recovery.
  }
  \label{fig:reward_velocity_grid}
\end{figure*}

\subsection{Behavior Analysis}
We quantify scenario complexity by the number of incoming vehicles present at the entrance of the roundabout. During training, the number of incoming vehicles $n$ is uniformly sampled from $\{0,1,2,3,4\}$ to expose the agent to a diverse set of driving situations. At test time, we evaluate the learned policy under three different complexity levels: low, medium and high corresponding to $n \in \{0,1,2\}$, $n = 3$ and $n = 4$, respectively. For every density we present the mean reward and velocity over 20 independent episodes.

Figures~\ref{fig:n2_rewards} and \ref{fig:n2_velocities} show each controller’s average reward and velocity profile with at most two interacting vehicles.
BC Transformer maintains an average reward above $0.9$, while CQL drops sharply owing to collisions in some of the $20$ evaluations.
 UWDT, DT, and SAC saturate at the maximum reward throughout the horizon. The velocity profiles in Figure~\ref{fig:n2_velocities} reveal the underlying behavior: BC Transformer cruises steadily at roughly $8\,\textrm{m/s}$; UWDT, DT, and SAC accelerate smoothly from $13$ to the $16\,\textrm{m/s}$ speed limit; while CQL's velocity remains low due to some early collisions. Because no vehicle must yield, all three high-performing agents merge without braking and achieve near-optimal performance.

With three interacting vehicle, the task becomes substantially harder. 
In Figure~\ref{fig:n3_rewards} the average rewards of BC Transformer and CQL drop sharply, indicating frequent collisions. 
SAC stabilizes around $0.8$, while UWDT and DT recover to a near-optimal reward after a brief transient. 
Figure~\ref{fig:n3_velocities} confirms that UWDT and DT reach the speed limit almost immediately and maintain it. 
SAC exhibits a full stop midway, waiting for a conservative gap before merging; once merged, it accelerates and finishes strongly. CQL decelerates to around $2.5\,\textrm{m/s}$ and seldom recovers, matching its poor reward performance.

At the highest traffic density, shown in Figure~\ref{fig:n4_rewards}, the BC Transformer deteriorates fastest, losing all reward within a few steps. CQL starts near $0.85$ but slips to $0.2$ by the end of the horizon as collisions accumulate. SAC’s mean episodic reward decreases to approximately $0.8$ and then shows a slight increase near the end.
The speed profile in Figure~\ref{fig:n4_velocities} indicates that SAC comes to a full stop and subsequently accelerates to $4.5\,\mathrm{m/s}$.
This pattern explains the reward decrement in SAC's reward profile. During the stop, the SAC agent loses speed-related reward while remaining collision-free. This conservative policy eliminates crash risk during roundabout merges, at the cost of driving efficiency. The SAC agent remains stationary even when sizable gaps emerge, degrading traffic throughput and increasing travel time. In our highest traffic density scenario, UWDT again dominates, sustaining nearly perfect rewards despite the crowded roundabout. Velocity traces in Figure~\ref{fig:n4_velocities} corroborate these findings. 
BC Transformer decelerates to rest almost instantly. UWDT accelerate briskly to the $16\,\textrm{m/s}$ limit and sustain that velocity. Across all densities, UWDT edges DT by incurring fewer collisions, yielding slightly higher cumulative rewards.

Overall, these results demonstrate that our UWDT consistently outperforms conventional RL algorithms and imitation learning baselines across varying traffic densities. By leveraging a transformer architecture with uncertainty‑weighted return conditioning, the UWDT maintains near‑optimal rewards and accelerates to the speed limit even in congested scenarios.

\section{conclusion}
\label{conclusion}
We developed and presented the Uncertainty Weighted Decision Transformer for autonomous navigation, which augments the standard transformer architecture with a principled mechanism for quantifying and exploiting epistemic uncertainty at inference time. Unlike purely behavior cloning baselines or conservative offline RL methods, UWDT conditions its action selection on both desired return and an uncertainty measure, allowing it to choose high‑reward trajectories when confident and adopt safer maneuvers when its predictions are less certain. Through extensive simulation experiments on a complex roundabout driving task, we demonstrated that UWDT achieves higher accumulated rewards, longer travel distances and shorter traversal times, and near‑optimal exit success rates while simultaneously reducing collision rates to the lowest value when compared to other tested methods
These results underscore the importance of incorporating uncertainty into sequential decision making, especially when critical situations are rare in the training set but have significant impact during deployment. UWDT improves safety and efficiency, making it a promising approach for safety critical driving applications. 

\section*{Acknowledgment}

\bibliographystyle{IEEEtran}
\bibliography{ref}

\vspace{12pt}

\end{document}